# A Video Database of Human Faces under Near Infra-Red Illumination for Human Computer Interaction Aplications

S L Happy, Anirban Dasgupta, Anjith George, and Aurobinda Routray

Department of Electrical Engineering

Indian Institute of Technology Kharagpur

*Abstract*—Human Computer Interaction (HCI) is an evolving area of research for coherent communication between computers and human beings. Some of the important applications of HCI as reported in literature are face detection, face pose estimation, face tracking and eye gaze estimation. Development of algorithms for these applications is an active field of research. However, availability of standard database to validate such algorithms is insufficient. This paper discusses the creation of such a database created under Near Infra-Red (NIR) illumination. NIR illumination has gained its popularity for night mode applications since prolonged exposure to Infra-Red (IR) lighting may lead to many health issues. The database contains NIR videos of 60 subjects in different head orientations and with different facial expressions, facial occlusions and illumination variation. This new database can be a very valuable resource for development and evaluation of algorithms on face detection, eye detection, head tracking, eye gaze tracking etc. in NIR lighting.

## I. INTRODUCTION

Human Computer Interaction (HCI) has been an active area of research for facilitating communication between computers and humans. It has been reported that face tracking [1], face pose estimation [2], face orientation [1], eye detection [3] and facial expression detection [4] have wide applications in HCI. There have been attempts [5], [6], [7], [8], [9] to develop face databases to test algorithms developed for such purposes. The databases referred in above literature use normal illumination using light of visible spectra range. However, certain applications such as biometric, surveillance and other military applications need communications with the system at night and under low illumination levels. In this paper, we describe the creation of a dataset recorded using NIR lighting.

Moreover, most of the existing databases for facial feature detection and person recognition incorporate frontal views only. However, this restricts the application area as well as constrains the subject's facial motion. In our database, the subjects were allowed to move the face in an unconstrained but systematic manner, to capture every possible orientation of the face. The database contains faces with different facial rotations as well as varying facial expressions. The subjects were from different geographical regions of India, which ensures diversity in the database. Another advantage of the database is that, being a video database, the runtime performance of any algorithm can be tested with the database along with accuracy analysis.

The main objectives of creating the NIR face database include the following:

- providing a standard platform for testing algorithms on face detection, face recognition, face tracking, head pose tracking, eye detection etc. in NIR lighting;
- testing the relative runtime performances of real-time algorithms on the video database; and
- providing a video database of faces from different parts of India.

This paper is organized as follows. Section II discusses the experiment design and steps followed while creating database. Section II describes about the photographic condition like camera and the lighting system and overall quality of collected data. Section IV contains the details about the database content. Section V highlights on the demographics of the database. In Section VI, the availability of the database is provided and section VII concludes the paper.

## II. EXPERIMENT DESIGN

An experiment was conducted in a laboratory in dark condition for six days from 6:00 pm to 9:00 pm with 10 subjects per day. The subjects were given video as well as verbal instructions regarding the tasks they had to perform. The recording time for each subject was eight minutes in average. The videos were captured using camera capable of capturing NIR radiation. An NIR lighting system made up of Gallium Arsenide LEDs is prepared taking into consideration the safety illumination limits. An audio file containing instructions for the subjects was played during the video recording. The instructions are listed in Table I.

TABLE I

INSTRUCTIONS FOR SUBJECTS

| | |
|---|---|
| Step 1. | Frontal face of subject with open and closed eyes. Subjects were asked to tell their name to introduce lip movement and facial changes. |
| Step 2. | Off-plane head rotation<br>a) Rotating face right and halting at 30°, 45°, 60° and complete right position<br>b) Moving face up and down while face is rotated<br>c) Repeating steps 2.a and 2.b with closed eyes<br>d) Repeating steps 2.a, 2.b and 2.c for left rotation |
| Step 3. | In-plane rotation (tilting)<br>a) Tilting head towards right and left<br>b) Moving face up and down after tilting<br>c) Repeating steps 3.a and 3.b for closed eyes |
| Step 4. | Up-down head movement<br>a) Movement of face up and down<br>b) Looking right and left while face is in up or down position |

This work is financially supported by Department of Information Technology, Government of India.

Download link: https://sites.google.com/site/nirdatabase/download

| | |
|---|---|
| | c) Repeating steps 4.a and 4.b with closed eyes |
| Step 5. | Rotation of head in a circular manner keeping face straight for both open and closed eyes |
| Step 6. | Diagonal movement of face<br>a) Moving face diagonally (at 45°) up-right, up-left, down-right, and down-left from the frontal position<br>b) Repeating step 6.a with closed eyes |
| Step 7. | Head nodding<br>a) Nodding head vertically<br>b) Nodding head horizontally<br>c) Nodding head randomly<br>d) Repeating steps 7.a, 7.b, 7.c and 7.d for closed eyes |
| Step 8. | a) Facial expressions showing basic emotions naming happiness, sadness, anger, disgust, surprise, and fear.<br>b) Apart from the basic emotions subjects were instructed to make some unusual expressions involving lip movement, eyebrow movement, facial muscle movement etc. |
| Step 9. | a) Subjects were instructed to occlude their face by putting hand on face or chick or by rubbing eye.<br>b) Head nodding was repeated while wearing spectacles. |
| Step 10. | Lighting variation was done manually. |

## III. DATA ACQUISITION

The database was created under NIR illumination, so radiations in the visible spectra were avoided. Hence, the experiment was conducted after switching off all light sources except the NIR illuminations. An arrangement was made to prevent the illumination from the laptop screen.

### A. Camera System

A USB camera having Charge Coupled Device (CCD) sensor and capable of capturing NIR illumination was used to record the videos. The recordings were performed at $640 \times 480$ pixels resolution at 30 frames per second (fps). Data from the camera is directly stored in computer through USB interface. A height-adjustable chair was used to keep the subject's face aligned with camera.

### B. Lighting System

The lighting system consisted of three arrays of Ga-As NIR LEDs. The LEDs were mounted on a movable stand which provided both horizontal as well as vertical movements. The power rating of the device was 3 W. The lighting intensity was varied by changing the relative position of the LEDs as well as changing the power supply. Fig. 6 shows some images of a subject with varying illumination.

### C. Backgrounds

The data was recorded in a complex background. But due to limited radiation range of the NIR lighting system, the background is totally dark. So our database basically contains face images with a dark background.

### D. Accessories

Apart from camera and NIR lighting system, a laptop was connected with USB camera to store the video. A stand is used to hold the camera as well as the NIR LEDs firmly.

### E. Discussion of Data Quality

In spite of the precautions taken and the adjustments made by the research team, noise persisted. Apart from this, some human errors were also introduced due to delayed response of the subject to the instructions and due to the misinterpretation and unfamiliarity of the subjects to certain instructions.

## IV. DATABASE CONTENT

The database consists of 60 subjects in various poses, expressions and occlusion conditions in NIR illumination. The subjects were from different parts of Indian subcontinent. Some of the men had beard/moustache. The subjects were in the age group of 20 – 40 years. There are 40 men and 20 women in total, and most of the subjects are of Indian origin.

### A. Head Poses

Most of the possible head poses have been acquired for each subject. Fig. 1 Off-plane head rotation at different anglesshows some sample head poses of a subject. The different face poses are obtained by instructing subjects by voice command to pose in a certain way. Voice instructions include off-plane face rotation in both directions, tilting of head, up-down movement and diagonal movement. For head tracking purpose circular head motion and head nodding are also included. Vertical, horizontal and random head nodding were also recorded as the part of the database. All the above mentioned instructions were repeated for closed eye case. Hence, the database captures a lot of variability in head movement and orientations.

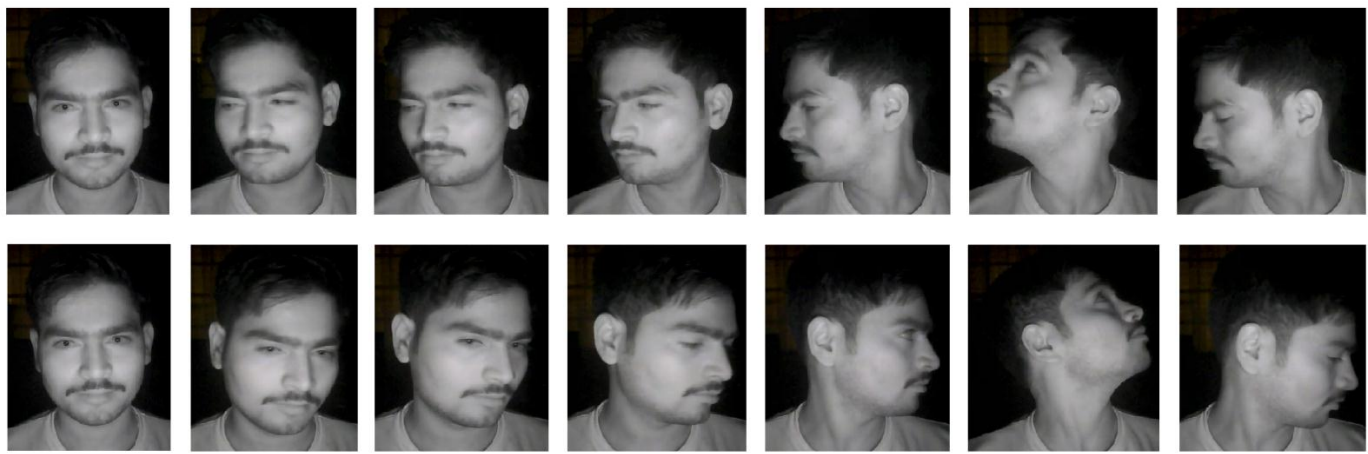

Fig. 1 Off-plane head rotation at different angles

*1) Off-plane rotation*

Subjects were instructed to move face right and left slowly using voice command and halting at 30°, 45°, 60° and 90°, hence, images of rotated face at different angles up to profile picture were obtained for both open and closed eye condition. Movement of face in up-down direction while rotating face has also been captured. Fig. 1 shows the off-plane rotation and moving face up and down while rotating left or right.

*2) In-plane rotation (Tilting)*

Subjects were instructed to tilt their head gradually in both right-left directions and then move face up and down subsequently as shown in Fig. 2.

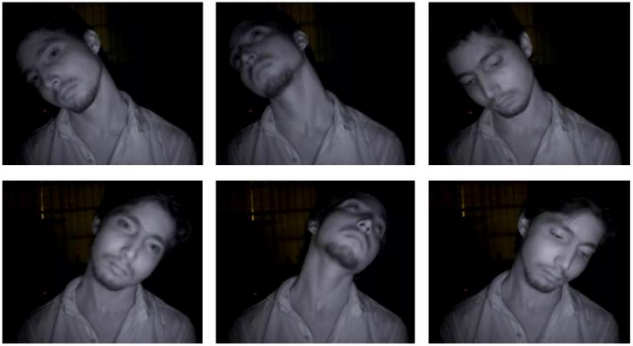

Fig. 2 In-plane head rotation, and up-down head movement after tilting

*3) Up-down orientation*

Up down movements of face were captured along with the face moving left and right while posing up and down which is shown in Fig. 3. This along with previous rotations ensures that all types of head positions have been captured.

*4) Diagonal movement of head*

Diagonal movement of head such as up-right, up-left, down-right, and down-left is also included in the database where subjects had to move face at an angle of approximately 45°. This will be helpful in tracking purpose while face has both in-plane and off-plane rotation.

*5) Head nodding*

Head nodding in vertical, horizontal and random direction are also recorded to add some more variation in head tracking.

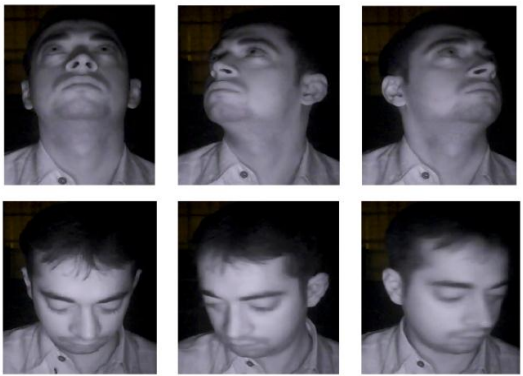

Fig. 3 Up and down face positioning with lateral movements

*B. Facial Expressions*

Apart from face rotations, the subjects were asked to exhibit different facial expressions based on different emotions. The subjects were asked to give their introduction to account for movement of lips. They were also asked to lift eyebrows up- down and fake eating something to bring more variations in the database. Fig. 4 shows some sample facial expressions from the database.

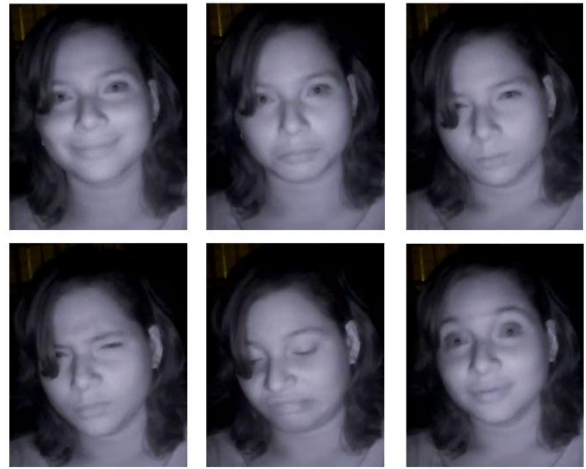

Fig. 4 Images of a Subject with different Facial Expressions

*C. Occlusions*

Most of the time face occlusions act as a barrier in face detection. To facilitate researchers with the NIR database containing occlusions, we have included natural occlusions like putting hand on face, cheek or rubbing eye. The subjects were also asked to wear powerless spectacles to occlude their eyes. Fig. 5 shows some occlusions of a subject's face during the experiment.

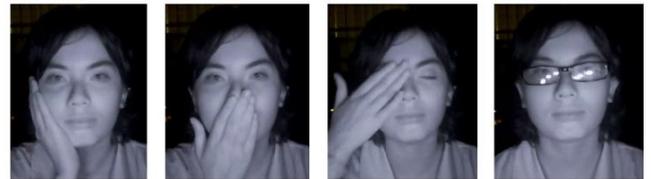

Fig. 5 Images of a Subject with occlusions

*D. Illumination variation*

The NIR lighting system was operated manually to illuminate face from different angles. By varying illumination levels face images were captured which is given by Fig. 6.

V. DEMOGRAPHICS OF THE DATABASE

Students and project staffs of IIT Kharagpur have volunteered to build this database. 60 subjects volunteered in total out of which 40 are male and 20 are female. Most participators are between the ages of 20 to 40 and are from different regions of India. Subject's details were determined by an optional questionnaire response.

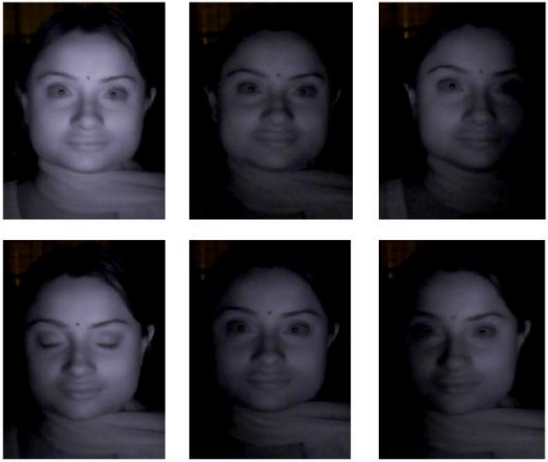

Fig. 6 Images of a Subject with varying illumination

## VI. AVAILABILITY

The database will be made available to interested parties by the authors upon request. One can also obtain the database by visiting https://sites.google.com/site/nirdatabase/download. A link will be provided containing the respective files. The facial images of some subjects may appear in research publications since they have granted permission in this regard.

## VII. CONCLUSION

In this paper, we have described the method of developing and contents of the database of facial images created under NIR lighting. The main characteristics of this face database are the variation, including head pose, expression, lighting, and the combined variations. The eye's open/close condition is considered as a special interest in the database. Researchers can use this highly diversified database with NIR illumination as a standard for training and testing purposes.

ACKNOWLEDGMENT

The authors would like to acknowledge the subjects for voluntary participation in the experiment for database creation.